\documentclass{vgtc}                          

\ifpdf
  \pdfoutput=1\relax                   
  \usepackage{graphicx}                
  \DeclareGraphicsExtensions{.pdf,.png,.jpg,.jpeg} 
\else
  \usepackage{graphicx}                
  \DeclareGraphicsExtensions{.eps}     
\fi%

\graphicspath{{figures/}{pictures/}{images/}{./}} 

\usepackage{microtype}                 
\PassOptionsToPackage{warn}{textcomp}  
\usepackage{textcomp}                  
\usepackage{mathptmx}                  
\usepackage{times}                     
\usepackage{cite}                      
\usepackage{tabu}                      
\usepackage{booktabs}                  

\onlineid{0}

\vgtccategory{Research}

\vgtcinsertpkg

\usepackage{caption}
\usepackage{subcaption}

\title{Time Series Model Attribution Visualizations as Explanations}

\author{Udo Schlegel\thanks{e-mail: u.schlegel@uni-konstanz.de} \\ %
     \scriptsize \centering University of Konstanz %
 \and Daniel A. Keim\thanks{e-mail: keim@uni-konstanz.de}\\ %
     \scriptsize University of Konstanz %
     }

\abstract{
Attributions are a common local explanation technique for deep learning models on single samples as they are easily extractable and demonstrate the relevance of input values.
In many cases, heatmaps visualize such attributions for samples, for instance, on images.
However, heatmaps are not always the ideal visualization to explain certain model decisions for other data types.
In this review, we focus on attribution visualizations for time series.
We collect attribution heatmap visualizations and some alternatives, discuss the advantages as well as disadvantages and give a short position towards future opportunities for attributions and explanations for time series.
} 

\CCScatlist{ 
  \CCScat{}{}{Human-centered computing}{Human computer interaction (HCI)};
  \CCScat{}{}{Computing methodologies}{Artificial intelligence}
}

\begin{document}


\firstsection{Introduction}

\maketitle

Explainable AI (XAI) introduces techniques as well as algorithms to open black-box models and support understanding, debugging, as well as refining of complex models~\cite{spinner_explainer_2019}.
Such techniques are essential to handle the growing democratization of deep learning models~\cite{hohman_visual_2018} and their state-of-the-art performance in an also increasing amount of research fields.
Nevertheless, critical application fields like healthcare or criminal justice need explanations to allow the usage of black-box models~\cite{rudin_stop_2019}.
Interpretable models are good baselines and alternatives in such critical cases.
However, as state-of-the-art performances are often only achieved by black-boxes, easily understandable explanations can overcome a few limitations.
Such explanations can help to combine peak performances with understandable decisions of complex models to tackle critical applications~\cite{ribeiro_why_2016}.
In most cases, XAI techniques, which open such deep learning black-box models, are compromised of local explanation using so-called attribution techniques~\cite{guidotti_survey_2018}.
Such attribution methods are developed to explain specific input samples for complex models, some model-agnostic~\cite{ribeiro_why_2016,lundberg_unified_2017} and a large amount model-specific~\cite{simonyan_deep_2013,bach_pixel_2015}.
Thus, various attribution techniques exist, and deciding which method to use is a rather tedious task as an evaluation is often either time-consuming with user preferences or computation heavy with automatic approaches~\cite{schlegel_towards_2019}.

The evaluation of XAI explanations is often divided into two categories: qualitative and quantitative evaluations~\cite{mohseni_survey_2018}.
The qualitative evaluation focuses on the human aspect and incorporates human understanding of explanations into the evaluation~\cite{ribeiro_why_2016}.
Quantitative evaluation is often done automatically, focusing on essential parts of the decisions in contrast to the input samples~\cite{hooker_benchmark_2019}.
A current trend focuses on the evaluation of attributions using automatic methods for images~\cite{hooker_benchmark_2019}, text~\cite{arras_evaluating_2019}, and time series~\cite{schlegel_empirical_2020} through perturbation analysis.
In such an evaluation analysis, the most important regions or parts of the input for the attribution are perturbed to a non-information holding value~\cite{zeiler_visualizing_2014}.
However, e.g., for time series such non-information values are not easily defined~\cite{schlegel_empirical_2020}.
Such automatic evaluations are often preferred as easily understandable visualizations for attributions are difficult to create for a general audience.
The current trend, e.g., in computer vision, presents heatmaps on the image input to visualize the attributions and relevances of corresponding pixels~\cite{guidotti_survey_2018}.
For instance, such heatmap visualizations can be transferred to time series but are rather hard to interpret even for experts~\cite{schlegel_towards_2019}.
Thus, the question arises, how can we support time series attribution visualizations to create understandable and robust explanations for changing demands and user groups.

This review collects and introduces attribution technique concepts for time series and presents visualization techniques for time series attributions.
We focus on approaches using heatmaps and collect related papers to present examples.
We highlight prominent approaches and discuss the advantages and disadvantages of such heatmap visualizations.
Further, we present first alternatives to such heatmaps for time series incorporating line plots.
At last, we introduce our position towards attribution visualization for different users and improved explanations as well as future opportunities for explanations on time series.

\section{Attributions on time series}
Attributions are local explanations of various XAI techniques and help to explain model decisions for single samples.
Global explanations present the overall decision-making of models and are hard to achieve for complex models such as deep neural networks.
In contrast, local explanations show only the decision-making for a single or a limited amount of input samples.
Attributions are a particular form of local explanations, which demonstrate the importance of input variables of a sample based on the predicted output of a model.
Thus, such attributions show how a model attributes output predictions to input features based on a single sample.

Attributions can essentially help to understand single decisions of a model towards one sample.
With different approaches, attribution methods generate a relevance score for every input variable of a sample using the input model as a base.
Collecting such relevance for the specific input then creates the attribution vector in the end.
In some cases, these attributions show the sensitivity towards specific input variable changes, e.g., occlusion~\cite{zeiler_visualizing_2014}. 
In contrast, in other approaches, additive attributions get calculated to give each feature value an additive score, e.g., SHAP~\cite{lundberg_unified_2017}.
There are many different approaches to generate such an attribution vector with other intentions and strategies. 
Nevertheless, each of these techniques assigns a score towards each input value into the model called attribution.

\textbf{In general} attributions can be generated for every input of every model as they show the importance of the feature on the selected sample regardless of the input dimensionality.
However, depending on the data type, other aspects are essential for attributions.
Image data has, in many cases, neighborhood importance for the pixels and regions with higher attributions.
On the other hand, tabular data does not incorporate neighborhoods as the features can be reordered without further challenges in many cases.
Correlations in tabular data are often not generated by the neighborhood, while pixels in images are highly correlated in the surrounding area.
Time series, in contrast, also form neighborhoods in the time direction while also adding correlations over features and time.
Revealing such correlations is not trivial and requires good attribution techniques with a way to communicate the explanation.
Thus, attributions can be calculated for complex time series models but are often difficult to interpret without visualizations or further abstract representations.

\begin{figure*}[ht]
     \centering
     \begin{subfigure}[b]{0.495\linewidth}
         \centering
         \includegraphics[width=0.95\linewidth]{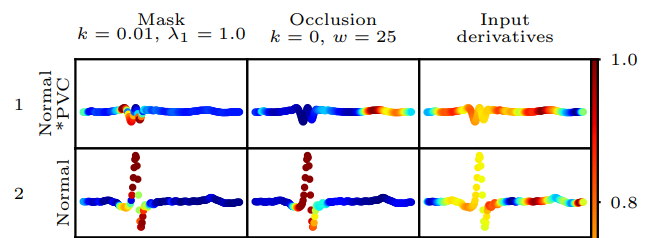}
         \caption{Van der Westhuizen and Lasenby~\cite{vanderwesthuizen_techniques_2017} visualize their feature importance of an LSTM model trained on healthcare data with the jet color scale (blue to green to red).}
         \label{fig:ts_xai_westhuizen}
     \end{subfigure}
     \hfill
     \begin{subfigure}[b]{0.495\linewidth}
         \centering
         \includegraphics[width=0.99\linewidth]{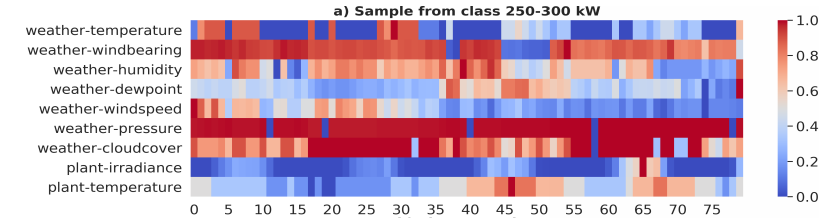}
         \caption{Assaf and Schumann~\cite{assaf_explainable_2019} visualize attributions for multi-variate time series as a heatmap using the attribution technique Grad-CAM with a color scale blue (small) to red (high).}
         \label{fig:ts_xai_assaf}
     \end{subfigure}
     \caption{Van der Westhuizen and Lasenby~\cite{vanderwesthuizen_techniques_2017} and Assaf and Schumann~\cite{assaf_explainable_2019} visualizations for multi-variate time series attributions as different heatmap approaches, directly on the line plot and as an abstract heatmap on the time points as rectangles. Van der Westhuizen and Lasenby~\cite{vanderwesthuizen_techniques_2017} visualize the feature attributions for ECG data with various features and an LSTM model. Assaf and Schumann~\cite{assaf_explainable_2019} present the feature attribution of Grad-CAM on different CNNs on energy consumption data on the individual time points and over time.}
\end{figure*}

A growing number of techniques can generate attributions, often categorized into gradient, structure, and surrogate techniques~\cite{samek_explainable_2017}.
These categories are based on the method they incorporate to extract attributions and are independent of the explanations.
Some of these methods can help to generate easier to understand explanations as they already implement some aggregations to enable abstractions, e.g., showing an attribution through addition on a baseline for the input dimensions~\cite{lundberg_unified_2017}.
Most others operate directly on the input to the model and present the raw attributions, which in some cases are not easily interpretable~\cite{jeyakumar_how_2020} or unreliable~\cite{kindermans_unreliability_2017}.

\textbf{Gradient methods} such as Saliency~\cite{simonyan_deep_2013} and GuidedBackpropagation~\cite{springenberg_striving_2014} use the gradient of the model to propagate the importance of specific output neurons back to the input neurons to generate attributions.
As gradients are fast to compute for single samples in most cases, gradient methods are thus a good starting point for attributions in general.
However, as gradients can be noisy due to the shattered gradients problem~\cite{balduzzi_shattered_2017}, gradient ensemble techniques such as SmoothGrad~\cite{smilkov_smoothgrad_2017} try to overcome these issues by adding noise to the input to improve attributions by smoothing the gradients over the noisy inputs.
Others such as Integrated Gradients~\cite{sundararajan_axiomatic_2017} incorporate a baseline input and slowly changing such a baseline to the selected input capturing the gradients and calculating the integral of the collected gradients to generate the attributions.

\textbf{Structure methods} such as LRP~\cite{bach_pixel_2015} and DeepTaylorDecomposition~\cite{montavon_explaining_2017} use the learned network weights and biases to propagate a score from the output to the input using specific weighting rules for each layer type.
Through such an auxiliary score, the challenges with computed gradients can be mitigated. 
However, other challenges arise, such as the selection of rules for the layers with, e.g.,  LRP~\cite{montavon_methods_2018} or stable references for, e.g., DeepLIFT~\cite{shrikumar_learning_2017}.
By providing a rule of thumb for the rule selection for the layers~\cite{montavon_layer_2019} and the references of the inputs, valuable attributions can be generated with both approaches.
Further, as the score is often straightforward to calculate even for deeper and wider models, these techniques are also relatively fast, enabling an easy comparison if the gradients are shattered, or other problems emerged with the gradient methods.

\textbf{Surrogate and sampling methods} such as LIME~\cite{ribeiro_why_2016} and SHAP~\cite{lundberg_unified_2017} first collect perturbed instances of the input sample and, based on these newly created data points, train an interpretable model or a game-theoretical model to generate the attribution scores.
One of the most considerable problems of these methods is the sampling, as better perturbed instances produce better interpretable models and thus better attributions~\cite{schlegel_ts_2021}.
Also, more perturbed samples lead to better attributions, but generating more perturbed instances takes more time as the model has to be applied to them for the prediction~\cite{ribeiro_why_2016}.
However, as these methods are model-agnostic and thus applicable to every input model and data, they are highly desired and needed by the industry for production-ready models.

\textbf{Evaluation} is essential for all attribution methods as the amount of possible applicable techniques is quite large, and the fidelity needs to be guaranteed.
In many cases, these methods are evaluated using either qualitative~\cite{ribeiro_why_2016} or quantitative~\cite{hooker_benchmark_2019, schlegel_towards_2019} methods.
Schlegel et al.~\cite{schlegel_towards_2019} present a quantitative evaluation with a fidelity perturbation analysis to find the most promising attribution techniques for time series, which focus on how accurate the explanations capture the model (trustworthiness).
Fig.~\ref{fig:ts_xai_technqiues} shows the comparison of a few evaluated approaches on uni-variate time series with a heatmap line plot visualization.
They perturb (exchange) time points with high relevance to a non-informative value.
As a non-informative value is challenging to define in time series, they use zero, the inverse, and the mean as such a value~\cite{schlegel_towards_2019}.
These fidelity perturbation approaches enable selecting the most fitting technique for the data type, and model architecture users can apply to time series~\cite{schlegel_empirical_2020}.
However, after detecting a suitable attribution technique, understandable communication as an explanation to users is still necessary as attributions hold relevances for every input variable for the given model.

\section{Heatmaps as basis for attribution explanations}
Attribution heatmaps are often used as explanation visualizations for images for attribution techniques.
Such a visualization technique presents the relevance value of the corresponding input variable on top of it.
Such an approach works quite nicely for images with a heatmap value right on top of the original pixel, and through opacity, it is still possible to combine both variables.
However, for instance, time series often use line plots for a baseline visualization, and such a combination of line plots and relevances leads to further challenges.

\begin{figure*}[ht]
     \centering
     \begin{subfigure}[b]{0.49\linewidth}
         \centering
         \includegraphics[width=0.99\linewidth]{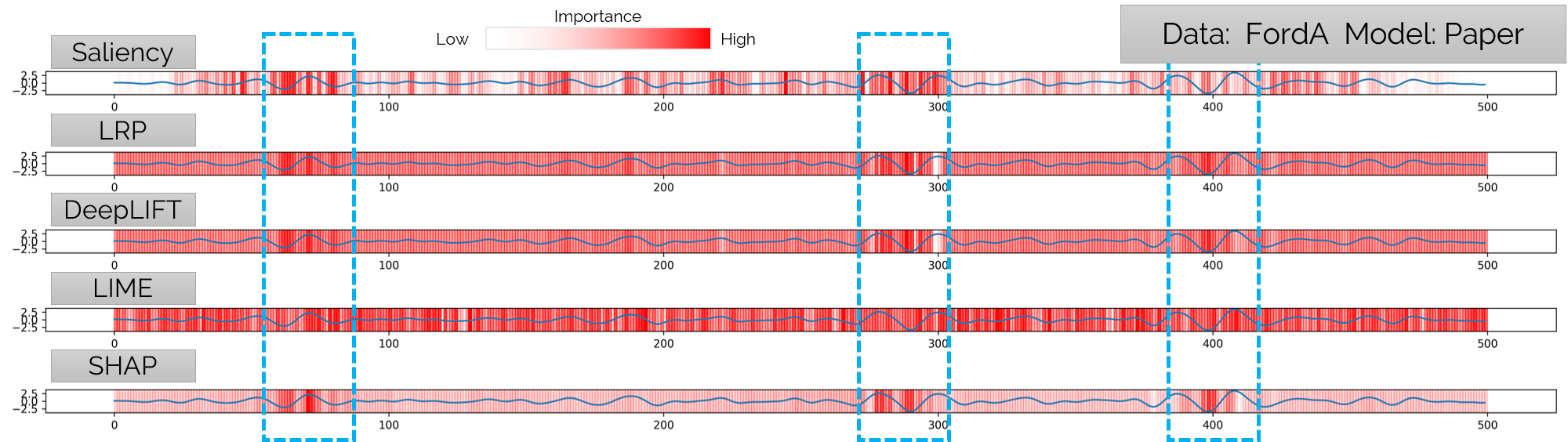}
         \caption{Schlegel et al.~\cite{schlegel_towards_2019} visualize attribution techniques for uni-variate time series as a heatmap from low (white) to high (red) in the background of a line plot.}
         \label{fig:ts_xai_technqiues}
     \end{subfigure}
     \hfill
     \begin{subfigure}[b]{0.49\linewidth}
         \centering
         \includegraphics[width=0.9\linewidth]{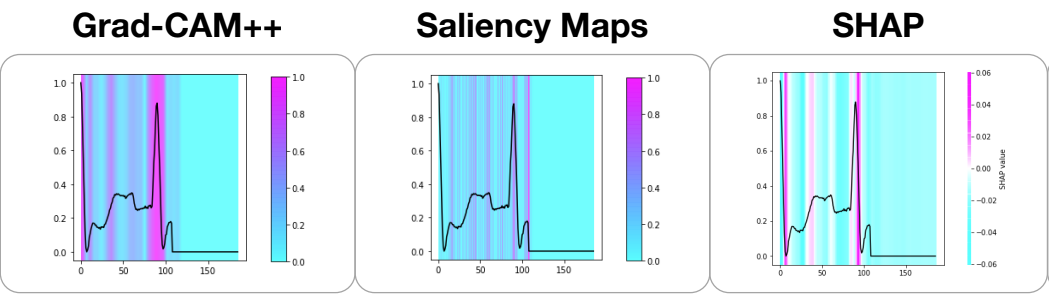}
         \caption{Jeyakumar et al.~\cite{jeyakumar_how_2020} create visualizations similar to Schlegel et al.~\cite{schlegel_towards_2019} with attributions in the background of the line plot ranging from low (cyan) to high (purple) values.}
         \label{fig:ts_xai_jeyakumar}
     \end{subfigure}
     \caption{Schlegel et al.~\cite{schlegel_towards_2019} and Jeyakumar et al.~\cite{jeyakumar_how_2020} visualizations for time series attributions focusing on multiple attribution techniques using heatmaps in the background to highlight the relevance. Schlegel et al.~\cite{schlegel_towards_2019} show the differences of attribution techniques and conduct a quantitative evaluation which of these attribution techniques are applicable and perform best on time series. Jeyakumar et al.~\cite{jeyakumar_how_2020} carry out a user study to compare the heatmap visualization of attribution techniques with other explanations such as explanation by example.}
\end{figure*}

Van der Westhuizen and Lasenby~\cite{vanderwesthuizen_techniques_2017} present how to show activations and attributions for LSTM networks on time series with saliency and occlusion.
They remove the line of a line plot and extend the size of the circles corresponding to the data with a color gradient at every time point to show the relevances of the attribution as seen in Fig.~\ref{fig:ts_xai_westhuizen}.
Through such a relatively small and direct approach to a heatmap on time series, challenges occur early. 
In their examples, it is hard to identify relevant time points due to the color gradient and, in some cases, overplotting. 
Especially without interaction methods, such a visualization leads to a distorted perception as later ones overdraw early rendered points.
However, for small-scale time series with a dynamic size of the circles, the visualization can lead to the first findings into the model.

Siddiqui et al.~\cite{siddiqui_tsviz_2019} also use a line plot as a baseline visualization.
They extend the line with a gradient towards relevant time points to show the attribution.
So, high attributions lead to a particular color (red) for specific parts of the line plot.
Thus, the overplotting issue and the generally distorted perception can be mitigated.
But, in some cases with bad gradients and coloring, the line plot itself can be rather hard to see, for instance, by having a gradient from white to red or distract the attention to mitigate intervals.

Assaf and Schumann~\cite{assaf_explainable_2019} remove the underlying data and show the attribution heatmap as a dense-pixel visualization by presenting each time point as a rectangle with the relevance score as the color as seen in Fig.~\ref{fig:ts_xai_assaf}.
By removing the time series data itself and only focusing on the attribution, they can investigate patterns and explain complex model decisions on time series.
However, due to limiting their attributions to Grad-CAM~\cite{selvaraju_grad_2017}, some found patterns are challenging to understand as they are even hard to explain with domain knowledge in the time domain by experts.

Viton et al.~\cite{viton_heatmaps_2020} take the same concept as Assaf and Schumann~\cite{assaf_explainable_2019} with another color scale going from light blue to orange and black as the zero value dividing these.
They exchange Grad-CAM~\cite{selvaraju_grad_2017} with some algorithm similar to the CAM approach but focusing more heavily on their neural network architecture.
Thus, they improve the attributions of their architecture. 
However, the visualization still limits the insights into the model itself and is in some cases due to the selected color scale harder to read than the others.

Schlegel et al.~\cite{schlegel_towards_2019} combine both approaches by visualizing a line plot of the time series with the attribution heatmap in the background as rectangles and the color of it as the relevance of the attribution.
Fig.~\ref{fig:ts_xai_technqiues} shows their approaches on various attribution techniques to highlight the contrasting explanations of the different methods.
Their approach tries to include the time series and attribution data, making the visualization harder to understand in general.
Especially, non-exerts need more time to understand the visualization as the line plots are relatively small while the heatmap is rather large. 
However, as the idea of the visualization is more about showing the differences of the attribution techniques, the focus shifts in general to experts.

Jeyakumar et al.~\cite{jeyakumar_how_2020} and Raghunath et al.\cite{raghunath_prediction_2020} use a similar approach to Schlegel et al.~\cite{schlegel_towards_2019} by presenting the attributions as a gradient behind the line plot (Fig.~\ref{fig:ts_xai_jeyakumar}).
Jeyakumar et al.~\cite{jeyakumar_how_2020} exchange the color gradient towards a more distinct scale from light blue to purple to highlight relevant regions.
Further, they conduct a user study to find out if such attribution visualizations help to understand the data or if nearest neighbor examples are better explanations.
Their visualization and also the one from Schlegel et al.~\cite{schlegel_towards_2019} demonstrate the problems of some of the attribution techniques as high relevance can be next to irrelevant time points.
Such challenges are hard to grasp for non-experts and experts alike as we expect the model to not learn specific time points by heart but some temporal patterns.
In some cases, these results are not consistent throughout the attribution techniques, which shows the problems of the attribution techniques on time series~\cite{schlegel_towards_2019}.

Heatmaps have some advantages and disadvantages for users in general.
For geospatial data, heatmaps help to encode more information for geo positions.
Thus, they help to present more information in an easy-to-understand way for most already intelligible environments (geo-maps or images in general).
However, there is a need for easier-to-understand visualizations, especially with time series and XAI explanations.
Showing the attention or focus of a black-box model helps identify specific characteristics but does not guarantee an understandable explanation for the model's decision-making.
In some cases in XAI and heatmap explanations, such explanations can be fooled and changed to some misguided or just wrong results~\cite{subramanya_fooling_2019,slack_how_2019}.
Thus, more abstract explanations are needed.


\section{Line plots with attribution extensions}
Possible alternative abstract representations for attributions are often grounded in line plots and their extensions for time series.
Temporal data has a long history of visualization techniques such as line plots, stacked plots, or horizon graphs~\cite{aigner_visualization_2011}.
However, especially adding more information to the temporal component visualization is not trivial.
For instance, multi-variate data is often visualized in small multiples and line plots to incorporate more information~\cite{aigner_visualization_2011}.
As we want to add more data into the same plot, such a visualization is even harder to compare and interpret for most user groups.

\begin{figure*}[ht]
     \centering
     \begin{subfigure}[b]{0.49\linewidth}
         \centering
         \includegraphics[width=0.99\linewidth]{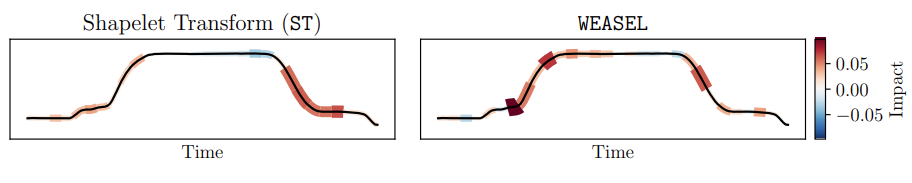}
         \caption{Mujkanovic et al.~\cite{mujkanovic_timexplain_2020} use pipes and a color gradient around a line of a line plot to visualize the relevance of an extended SHAP algorithm for time series.}
         \label{fig:ts_xai_timexplain}
     \end{subfigure}
     \hfill
     \begin{subfigure}[b]{0.49\linewidth}
         \centering
         \includegraphics[width=0.9\linewidth]{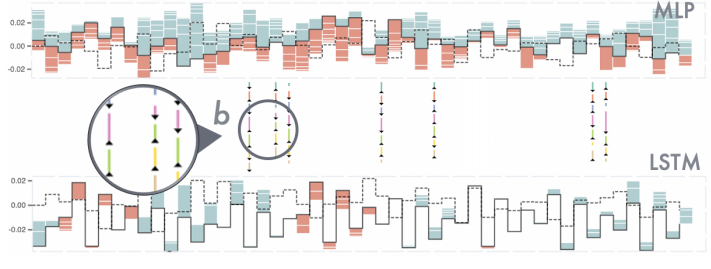}
         \caption{Xu et al.~\cite{xu_mtseer_2021} show the additive attributions of SHAP~\cite{lundberg_unified_2017} as bar charts on the forecast target and further arrows to compare multiple models attributions.}
         \label{fig:ts_xai_xu}
     \end{subfigure}
     \caption{Mujkanovic et al.~\cite{mujkanovic_timexplain_2020} and Xu et al.~\cite{xu_mtseer_2021} visualizations for time series attributions focusing on either a pipe enhancement for line plots to visualize relevance of SHAP or multiple models using the additive attribution SHAP technique for an improved comparison between various models. Mujkanovic et al.~\cite{mujkanovic_timexplain_2020} extend the SHAP technique for time series and visualize the corresponding relevance as pipes around the line plot with a color scale. Xu et al.~\cite{xu_mtseer_2021} incorporate the additive properties of SHAP to compare the performances of time series forecasting models by enabling explanations of single models and a direct comparison with arrows showing the differences in the attributions.}
\end{figure*}

Siddiqui et al.~\cite{siddiqui_tsinsight_2020} propose with TSinsight an attribution technique but also use multiple line plots to show the attribution as a line plot.
Thus, they move the visualization towards small multiples in the form of multiple line plots.
Their visualizations show the relevances of time series in another plot which enables a closer look at the initial data and the attributions to, e.g., compare different time points and their relevance more easily. 
However, the relation between time series and attribution data is more complex because locating the correct time point in the time series data is not trivial by focusing on another attribution line plot.
Their focus moves from the combination of the time series and the attributions towards the attributions, which can help, for example, experts with domain knowledge of certain situations, e.g., time points on anomaly events.

Mujkanovic et al.~\cite{mujkanovic_timexplain_2020} present another technique to generate attributions and an approach towards combining line plots with heatmaps.
However, these heatmaps are not just a color gradient on or behind the line of a line plot but mapped on the color and size of a region around the line.
They use a pipe around the line to present the relevance of the attributions (Fig.~\ref{fig:ts_xai_timexplain}).
Such an approach has two advantages; on the one hand, the focus of the user switches to the larger pipes as the size and the color highlight the part of the time series, which can potentially have the highest relevance for the prediction.
On the other hand, as little relevance can still be interesting, the color and a small pipe enable users to still find and see the less important regions and parts of the time series.

Line plots are generally the first visualization for temporal data~\cite{aigner_visualization_2011}.
However, in many cases, the temporal data is the only visualized component without further additional information towards, e.g., explanations.
Combining both is challenging and often done with color gradients as heatmaps on the line or behind it in the line plot.
However, such heatmaps line plots approaches are either difficult to understand for non-experts or reveal the issues of attribution techniques showing high next to low relevances.
Some of these challenges can get mitigated by easy-to-understand visualizations on line plots with aggregations such as Mujkanovic et al.~\cite{mujkanovic_timexplain_2020}.


\section{Future Opportunities}

Another approach by Xu et al.~\cite{xu_mtseer_2021} incorporates the additive properties of SHAP~\cite{lundberg_unified_2017} into their visualization to exchange the heatmap with bar charts on top and beneath a line representing the time series data seen in Fig.~\ref{fig:ts_xai_xu}.
Such an approach enables overcoming attribution heatmaps with more data to allow experts to investigate the decisions of models in more detail.
Significantly, such an approach helps present different features' influences in a multi-variate time series on the forecasting target.
The bar charts above and beneath the line plot show the additive influences of the SHAP~\cite{lundberg_unified_2017} attributions enabling a direct comparison of the features and also even models.
Fig.~\ref{fig:ts_xai_xu} shows their approach on a Multi-Layer Perceptron (MLP) and a LSTM~\cite{hochreiter_long_1997} model with a more widespread attribution for the MLP.
They even introduce a comparison visualization to compare the attributions shown in (b) with arrows and color lines to encode the differences in direction and value.
Their visualization lacks an application towards uni-variate time series as the additive property is not given, but such an approach works nicely for multi-variate time series and multiple models.

In general, such visualizations are hard to understand for non-experts and often also for experts~\cite{jeyakumar_how_2020}.
Jeyakumar et al.~\cite{jeyakumar_how_2020} compare heatmap visualizations of attributions against explanations by example.
Their user study clearly demonstrates that explanations by example are better suited for time series than heatmaps of Grad-CAM++~\cite{chattopadhay_grad_2018}, Saliency~\cite{simonyan_deep_2013}, and SHAP~\cite{lundberg_unified_2017} on time series data.
Thus, especially for non-experts, other explanations than attribution heatmap visualizations are preferable and advisable. 
We argue for incorporating counterfactuals as explanations for time series to non-experts but also for experts as these and contrastive explanations are more robustly grounded in human decision making and their explanations~\cite{byrne_counterfactuals_2019}.
Counterfactuals are instances of an input sample that flip the predictions of the input sample by changing it only marginally.
With the help of the Shneiderman Mantra~\cite{shneiderman_eyes_1996}, we propose the following approach for time series explanations.
As a first step, counterfactuals demonstrate the first glance as an explanation and give an overview of the model's internal decisions on specific samples.
Thus, the original sample is visualized with corresponding counterfactuals visualizations nearby to the sample to enable a direct comparison either in the data domain or some projection of it.
In the next step, an in-depth analysis can be supported by attribution visualizations by moving to the data domain level with a combination of time series and attribution data.
By providing further interaction techniques, e.g., a what-if analysis, a probing of a user towards the selected sample, and individual time points, enables further investigation into the details of the model.
For instance, with counterfactuals and assisted probing of time points, an in-depth analysis of decision boundaries of critical time points is possible to mitigate attacks and analyze the robustness of the model.
Thus, the approach focuses on overview first (counterfactuals), zoom, and filter (individual attributions), with details on demand (attributions and time series data interaction).
Such an approach includes the advantages of attributions and counterfactuals while mitigating many disadvantages of, e.g., attributions with additional information and limited heatmap visualizations as explanations.


\section{Conclusion}
We introduced attribution technique concepts for time series and their most common visualization technique (heatmaps).
We further collected and presented related work for attribution visualizations for time series models based on these concepts.
By analyzing these works, we argue for other visualization options such as enhanced line plots and argue for more abstract visualizations moving away from heatmaps and line plots to present explanations better for attributions.
We argue for counterfactuals instead of attributions for time series for non-experts and attributions only for domain experts.
At last, we present a small-scale workflow to combine counterfactuals and attributions into one pipeline approach to support non-experts and experts in their analysis of black-box time series models.

\acknowledgments{
This work has received funding from the European Union’s Horizon 2020 research and innovation programme under grant agreement No. 826494.}

\bibliographystyle{abbrv-doi}

\bibliography{template}
\end{document}